\documentclass[11pt]{article}
\usepackage{times}
\usepackage{url}
\usepackage{latexsym}
\usepackage{arabtex}
\usepackage{utf8}
\usepackage{multirow}
\usepackage{amsmath,amsfonts,amssymb}
\usepackage{mathtools}
\usepackage{graphicx}
\usepackage{subcaption}
\usepackage{svg}
\usepackage[T1]{fontenc}

\usepackage{siunitx}
\usepackage{authblk}
\usepackage{coling2020}
\usepackage{hyperref}

\DeclarePairedDelimiter\abs{\lvert}{\rvert}%

\hypersetup{
    colorlinks=true,
    linkcolor=red,
    citecolor=blue,
    filecolor=magenta,
    urlcolor=cyan,
}

\begin{document}
\title{Deep Diacritization: Efficient Hierarchical Recurrence for Improved Arabic Diacritization}

\author[1]{Badr AlKhamissi}
\author[1,2]{Muhammad N. ElNokrashy}
\author[2]{Mohamed Gabr}

\affil[ ]{
    {\tt
    \href{mailto:badr@khamissi.com}{badr@khamissi.com},
    \href{mailto:muhammad.nael@gmail.com}{muhammad.nael@gmail.com},
    \href{mailto:mohamed.gabr@hotmail.com}{mohamed.gabr@hotmail.com}}
}

\affil[1]{The American University in Cairo (AUC)}

\affil[2]{Microsoft Egypt Development Center (EGDC)
}

\renewcommand\Affilfont{\fontsize{10}{10}\itshape}

\setcode{utf8}
\maketitle

\begin{abstract}
    We propose a novel architecture for labelling character sequences that achieves state-of-the-art results on the Tashkeela Arabic diacritization benchmark. The core is a two-level recurrence hierarchy that operates on the word and character levels separately---enabling faster training and inference than comparable traditional models. A cross-level attention module further connects the two, and opens the door for network interpretability. The task module is a softmax classifier that enumerates valid combinations of diacritics. This architecture can be extended with a recurrent decoder that optionally accepts priors from partially diacritized text, which improves results. We employ extra tricks such as sentence dropout and majority voting to further boost the final result. Our best model achieves a WER of 5.34\%, outperforming the previous state-of-the-art with a 30.56\% relative error reduction.
\end{abstract}

\section{Introduction}
\label{intro}

%
%
\blfootnote{
    %
    %
    %
    %
    \hspace{-0.65cm}  
    This work is licensed under a Creative Commons 
    Attribution 4.0 International Licence.
    Licence details:
    \url{http://creativecommons.org/licenses/by/4.0/}.
    %
    %
}
\blfootnote{\hspace{-0.65cm}
    Affiliation emails:
    \fontsize{8.5}{9}
    {\tt \{\href{mailto:balkhamissi@aucegypt.edu}{balkhamissi}, \href{mailto:m.n.elnokrashy@aucegypt.edu}{m.n.elnokrashy}\}@aucegypt.edu};
    {\tt \{\href{mailto:muelnokr@microsoft.com}{muelnokr},
      \href{mailto:mogabr@microsoft.com}{mogabr}\}@microsoft.com}
}
\blfootnote{\hspace{-0.25cm}
    *\,This work is not sponsored by the affiliated institutions of the authors.}
\blfootnote{\hspace{-0.25cm}
    **\,This work was accepted at the Fifth Arabic Natural Language Processing Workshop (WANLP 2020).}

The Arabic script (and similarly Hebrew, Aramaic, Pahlavi...) is an impure abjad. These writing systems represent short consonants and long vowels using full letter graphemes, but generally omit short vowels and consonant length from writing. This leaves the task of inferring the missing phonemes to the reader by using context from neighbouring words and knowledge of the language structure to determine the correct pronunciation and disambiguate the meaning of the text. Those sounds are represented by diacritical marks---small graphemes that appear usually above or below a basic letter in the abjad. Table \ref{diac-table} shows the diacritics considered in this work. Diacritics are usually utilized in specific domains where it is important to explicitly clear up ambiguities or where inferring the correct forms might be difficult for non-experts, such as religious texts, some literary works such as poetry, and language teaching books as novice readers have yet to build up the intuition for reading undiacritized text.

We focus in this work on diacritization of Arabic texts. However, our proposed architecture has no explicitly language-dependent components and should be adaptable for other character sequence labelling tasks. Although it is the first language of several million people, and is spoken in some of the fastest growing markets \cite{tinsley_board}, the Arabic language, like many others, lacks attention from the NLP community compared to established test bed languages such as English or Chinese, which both enjoy higher momentum and an abundance of established resources and techniques. The automatic restoration of diacritics to Arabic text is arguably one of the most important NLP tasks for the Arabic language. Besides direct applications like facilitating learning, diacritics are used to enhance language modeling, acoustic modeling for speech recognition, morphological analysis, machine translation, and text-to-speech systems (which need to restore the lost phonemes to render words properly) \cite{zitouni09,azmi_2013}.

To illustrate this further, Table \ref{elm-table} shows the Arabic word \textit{Elm}\footnote{This paper uses Buckwalter transliteration \label{foot:buckwalter}} in different diacritized forms with their corresponding English translations, showcasing the importance of diacritics in resolving ambiguity. Note that the MADA \cite{MADA} morphological analyzer produces at least 13 different forms for this undiacritized word \cite{belinkov-glass-2015-arabic}.

\begin{table}[ht]
\centering
\begin{tabular}{|r|l|l|}
\hline 
\bf Arabic (diacritized) & \bf Transliteration & \bf English Translation \\ \hline
\<عَلِمَ>   &   \textit{Ealima}          & He knew \\
\<عُلِمَ>   &   \textit{Eulima}          & It was known \\
\<عَلَّمَ>   &   \textit{Eal$\sim$ama}    & He taught \\
\<عِلْمُ>   &   \textit{Eilomu}          & Knowledge \\
\<عَلَمُ>   &   \textit{Ealamu}          & Flag \\
\hline
\end{tabular}
\caption{Subset of possible diacritized forms for \textit{Elm} adapted from \protect\cite{belinkov-glass-2015-arabic} }
\label{elm-table}
\end{table}

Table \ref{diac-table} shows the different diacritics commonly used in Arabic texts along with their phonemic symbols. They fit roughly into four kinds. (1) \d{H}arakāt are diacritics for short vowels; we have three: fat\d{h}ah, kasrah, dammah. The symbols for those vowels have another form (usually a visual doubling) used at the end of a word to form a (2) tanwīn, or nunation, which is a VC sound of the \d{h}arakah's vowel followed by the consonant ``n'' ($\{a,i,u\}n$). (3) The shaddah is the gemination symbol used to indicate consonant doubling. It can be combined with one of the \d{h}arakāt or tanwīn on the same character. Finally, (4) the sukūn is used to indicate that the current consonant is not followed by a vowel and instead forms a cluster with the next consonant. Diacritics which appear at the end of a word are referred to as case-endings (CE); most of which are specified by the syntactic role of the word. They are harder to infer than the core-word diacritics (CW) that specify lexical selection and appear on the rest of the word \cite{mubarak19-highly}.

\begin{table}[ht]
\begin{center}
\begin{tabular}{|r|l|l|l|l|}
\hline 
\bf Symbol & \bf Name & \bf Type & \bf Transliteration & \bf IPA phoneme \\ \hline
\<هُ> & dammah       & \d{h}arakāt   & \it u & /u/ \\
\<هَ> & fathah       & \d{h}arakāt   & \it a & /a/ \\
\<هِ> & kasrah       & \d{h}arakāt   & \it i & /i/ \\
\<هٌ> & dammatain    & tanwīn        & \it N & /un/ \\
\<هً> & fathatain    & tanwīn        & \it F & /an/ \\
\<هٍ> & kasratain    & tanwīn        & \it K & /in/ \\
\<هّ> & shaddah      & shaddah       & $\sim$ &  /h:/ Gemination. \\
\<هْ> & sukūn        & sukūn         & \it o & No vowel. \\
\hline
\end{tabular}
\end{center}
\caption{Primary Arabic diacritics on letter \<ه>}
\label{diac-table}
\end{table}

The paper is structured as follows: First we cover some of the approaches used in related works on restoring Arabic diacritics. Then we introduce our system and support it by comparing experimental results on an adapted version of the Tashkeela corpus \cite{tashkeela17} proposed by \cite{fadel19} as a standard benchmark for Arabic diacritization systems. Each design decision will then be motivated by an ablation study. We analyze the learned attention model then discuss existing limitations in an error analysis. Finally, we offer directions for future work.

\section{Related Work}

The literature surrounding the automatic diacritization of Arabic text provides methods in two categories: classical rule-based solutions, and statistical modeling-based methods. Early approaches have worked on constructing a large set of language specific rules to restore the lost diacritics \cite{MADA,zitouni06,pasha14-madamira,darwish17}. Researchers have then shifted to rely more on learning-based methods that do not require extra expert systems such as morphological analyzers and part-of-speech taggers. \cite{belinkov-glass-2015-arabic} have shown that recurrent neural networks are suitable candidate models for learning the task entirely from data and can be easily extended to other languages and dialects without the use of manually engineered features. Other methods such as hidden Markov models (HMMs) \cite{elshafei06}, conditional random fields (CRFs) \cite{darwish17}, maximum-entropy models \cite{zitouni06} and finite-state transducers \cite{shieber05} have similarly been employed. However, more recent works have started to use deep (neural-based) architectures such as sequence-to-sequence transformers and recurrent cell-based models inspired by work in Neural Machine Translation \cite{mubarak19-highly}. Solutions combining both rule-based and deep learning methods appear in recently published work \cite{hamza20}. \cite{joint2020} have shown that the diacritization task benefits from jointly modelling lexicalized and non-lexicalized morphological features instead of targeting only the diacritization task.

\section{Approach}

\subsection{Datasets}
We report on the cleaned version of the Tashkeela corpus \cite{fadel19}---a high quality, publicly available dataset. It is split into train (\si{2,\!449 k} tokens), dev (\si{119 k} tokens), and test (\si{125 k} tokens) sets.

\subsection{Architecture}
\label{sub:arch}

\begin{figure}[ht]
    \centering
    \includegraphics[width=1\linewidth]{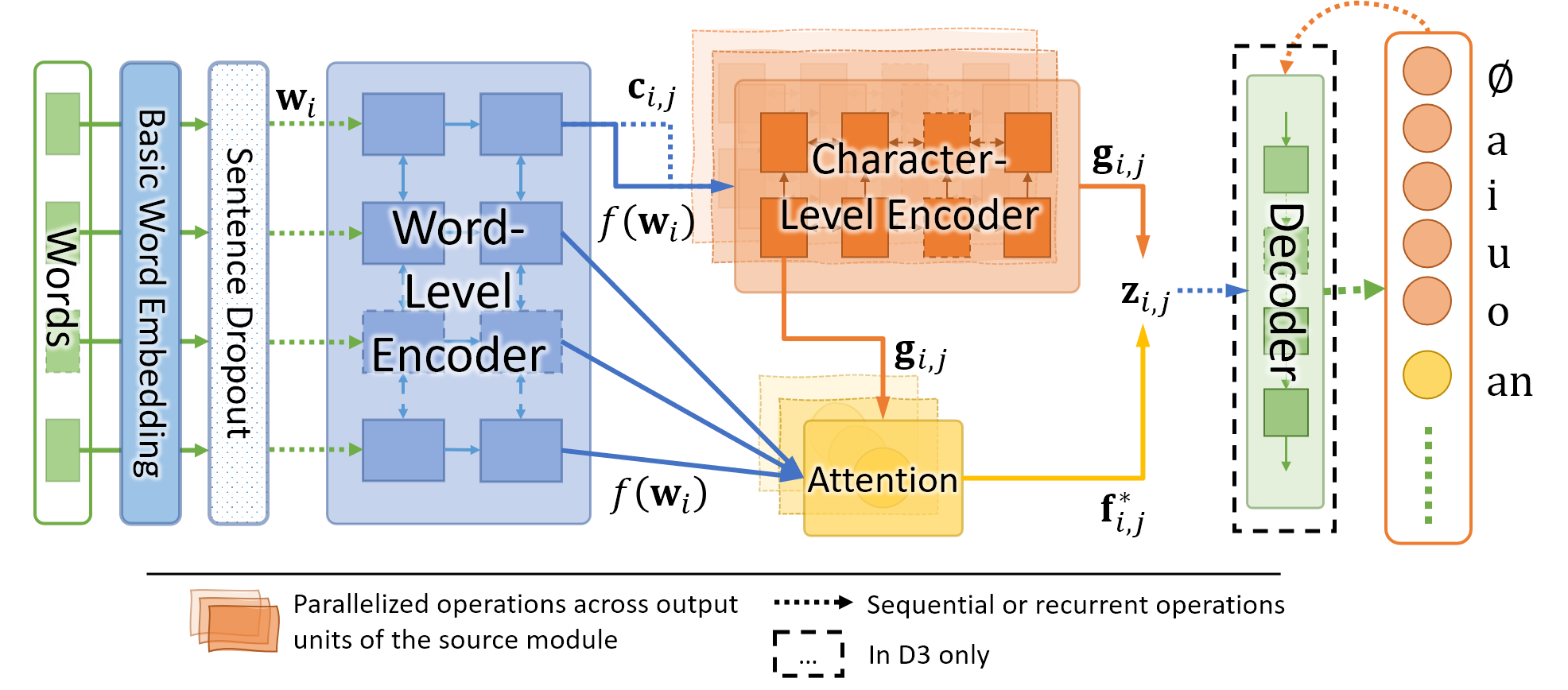}
    \caption{D3 Architecture}
    \label{fig:arch}
\end{figure}

In this work, we propose two models: The Two-Level Diacritizer (D2) and the Two-Level Diactritizer with Decoder (D3). D3 extends D2 by allowing partially diacritized text to be taken as input---a much needed feature for neural diacritizers \cite{fadel19}. D2 outperforms fully character-based models in both task and runtime performance measures.
\subsubsection{Two-Level (Hierarchical) Model}
\label{arch:hie}
Restoring diacritics can be seen as a character sequence labeling task. Each label depends on word- and character-level context. This structure motivates the hierarchy in our two-level encoder architecture---the first encoder sees the sequence of words (where words are atoms) and provides word-level context for the second, character-level encoder. The character-level encoder is evaluated independently for each word in the sentence, enabling much faster training and inference compared to character-level recurrent cell-based models of similar structure to previous works \cite{belinkov-glass-2015-arabic,mubarak19-highly,joint2020,fadel19-neural}. Let $T_s$ be the maximum number of words allowed in a sentence and $T_w$ the maximum number of characters allowed in a word. Then the overall character sequence length that a model in prior works would see is in the order of $(T_s \cdot T_w)$. In contrast, our approach operates on a maximum sequence length of $T_s$ at the word level and $T_w$ at the character level. Because character-level recurrence is independent of characters outside the current word, the serial bottleneck complexity goes from $O\left(\sum_{w \in s}{\abs{w}}\right) \approx O(T_s \cdot T_w)$ down to $O\bigl(\abs{s} + \max_{w \in s}\{\abs{w}\}\bigr) \approx O(T_s + T_w)$, assuming adequate parallelization across word units. See Table \ref{speed-table} for a speed comparison.

Let $s = \{w_i\}_{i=1}^{T_s}$ denote a sequence of words. $\mathbf{w}=\{\mathbf{w}_i\}_{i=1}^{T_s}$ are the corresponding fastText features pretrained on CommonCrawl Arabic data \cite{fastText17}. Let $w_i = \{c_{i,j}\}_{j=1}^{T_w}$ denote the sequence of characters for the word at index $i$ in sentence $s$. Each character is assigned a 32-dimensional learned embedding $\mathbf{c}_{i,j}$. Let $f(\mathbf{w})_i$ denote the feature vector from the word-level encoder which maps each word in $s$ to a contextual word representation. Let $g(\cdot)_{i,j}$ denote the character-level recurrence that
outputs a contextual encoding of the character relative to its parent word and sentence. See Figure \ref{fig:arch} for an overview. Formally, the contextual embedding $\mathbf{z}_{i,j}$ of $c_{i,j}$ is
\begin{align}
    \mathbf{g}_{i,j} &=
    g\left(
        \left[
            \mathbf{c}_{i,j} \, ; \,f(\mathbf{w})_i
        \right]
    \right)
    \\
    \mathbf{z}_{i,j}
        &=\left[
            \mathbf{g}_{i,j}
            \,;\,
            \mathbf{f}^*_{i,j}
        \right]
\end{align}
Where $\mathbf{f}^*_{i,j}$ is the attention view. Both $f(\cdot)$ and $g(\cdot)$ use Bidirectional LSTM (Bi-LSTM) layers \cite{graves05} trained with backpropagation through time. We note that any similar sequence modelling architecture would be applicable \cite{gru14,attention} but leave that to future work.

\subsubsection{Cross-level Attention Module}
\label{arch:attn}
This module attends over the word-level encodings $f(\mathbf{w})$ based on the character encodings $g(\cdot)$ of each character in a word. In other words, it uses the initial contextualization of the characters ($\mathbf{g}_{i,j}$) to attend to all words in the sentence (except the current) to refine the character's representation. We use the attention formulation from \cite{attention}. For each character $c_{i,j}$, we calculate 

\begin{align}
    \mathbf{f}^*_{i,j}
    &= \operatorname{AttendReduce}\left(
        \mathbf{u} = \mathbf{g}_{i,j}
        \,;
        \mathbf{X} =
        \{... f(\mathbf{w})_{0 : i-1} , f(\mathbf{w})_{i+1 : T_s} ...\}
    \right)
\end{align}

where
\begin{align}
    \label{eq:Attention}
    \operatorname{AttendReduce}\left(\mathbf{u}; \mathbf{X}\right)
        &= W^O \left(
            \operatorname*{Softmax}_t \left(
            \frac
            {
                W^Q(\mathbf{u})
                \cdot
                W^K(\mathbf{X})_t
                ^\top
            }
            {\sqrt{d_K}}
        \right)
            \cdot
            W^V\!\left(\mathbf{X}\right)
        \right)
\end{align}

 where in eq. \eqref{eq:Attention}, $W^Q$, $W^K$, $W^V$, and $W^O$ are independent linear layers. We tried to remove $W^O$ but faced lower performance.

\subsubsection{Decoder}
Used in D3, this component is a forward-only LSTM that takes as input a concatenation of the basic contextual character embedding $\textbf{z}_{i,j}$ and a one-hot representation of the output of the previous character from the classifier module. Formally: $[\textbf{z}_{i,j} ; \hat{\textbf{y}}_{i, j-1}]$. The $\hat{\mathbf{y}}_{i, j-1}$ signal passed to the decoder also encodes the Beginning-of-Word in addition to the previous-character diacritics. This allows the model to accept partially diacritized sentences such that the ground truth diacritic is injected in place of $\hat{\textbf{y}}_{i, j-1}$ during inference. This feature is important as many Arabic texts contain sparse diacritics that act as hints to assist readers. Having this clean signal yields improvements as shown in Figure \ref{fig:pd}.

\subsubsection{Task Objective}
The final classifier optimizes a Softmax objective over an enumeration of all valid diacritic combinations. Combined, we have 3 \d{h}arakāt in 4 variants (with tanwīn, shadda, tanwīn and shadda, and neither), the sukūn, and the plain shadda. Thus 15 classes including the None (no diacritic) class.

\subsection{Experimental Setup}

\subsubsection{Parameters, Hyper-parameters, and Regularization}

\paragraph{Optimization}
We use the Adam optimizer \cite{adam14} with an initial learning rate of $0.002$. The model is left to converge until the validation loss does not improve for $3$ consecutive epochs where each epoch enumerates a randomly shuffled version of the training segments exactly once. The learning rate is reduced by half when the validation loss does not improve for one epoch. We train with a mini-batch size of $128$ segments.

\paragraph{Encoders and Decoder}
\label{exp:enc-dec}
The word and character level encoders are \emph{each} a 2-layer stacked Bi-LSTM with $256$ and $512$ hidden units, respectively. We apply feature-level dropout \cite{dropout} with probability $0.2$ to the input of the character level encoder. The decoder in D3 is a one-layer forward-only LSTM with $1024$ hidden units. All recurrent cells use a vertical and recurrent dropout of $0.25$ each. The recurrent dropout used is untied between time-steps, in contrast to \cite{rec-dropout}.

\paragraph{Context Window and Voting}
\label{exp:voting}
Similar to \cite{mubarak19-highly}, we use a sliding context window of size $T_s$ on each sentence. A given sentence is split into several overlapping segments each of which is given separately to the model during training. This works well as the local context is often sufficient for correct inference. During inference, the same sequence of characters may appear in different contexts (different segments from one sentence) and potentially lead to different diacritized forms. To choose the final diacritic, we use a popularity voting mechanism and, in the case of a tie, choose one of the outputs at random. The values chosen for $T_s$, $T_w$ and the \textit{stride} are: $T_s=10$ with $\textit{stride} = 1$ for training and validation (a small $T_s$ was observed to stabilize training and improve results); $T_s=20$ with $\textit{stride} = 2$ for evaluation/testing; and $T_w=13$ for both training and evaluation.

\paragraph{Sentence Dropout}
\label{exp:sdo}
We randomly dropout 20\% of the words given to the word-level encoder during training. The positions of the dropped out words are preserved, and their embedding vectors $\mathbf{w}_i$ are replaced with zeros. This was observed to lead to better generalization in some cases.

\subsubsection{D3 Training}
\label{d3-training}
This model is not trained from scratch, but uses the weights of the encoders and character embeddings learned from D2. Those weights are kept frozen and only the decoder and classifier are trained.

\paragraph{Ramp-up of Teacher-forcing Signal}
\label{exp:gt-rampup}
We pass the ground truth of $p\%$ of the previous-character diacritics as input to the decoder at the current time-step. This value is ramped up from $p=0\%$ (all characters receive previous diacritics as zeros; i.e. no signal) to $p=100\%$ (all characters receive ground truth of previous diacritic as signal). This is done over a period of $n=10$ epochs in increments of $10\%$. Then the model is left to converge using the same stopping criteria as in D2. We found this to be the best approach as otherwise the model overfits early on the teacher forcing signal given from the previous time-step. 

\subsubsection{Source Code}
The code is made open source and is available on GitHub\footnote{\url{https://github.com/bkhmsi/deep-diacritization}}. We also provide an accompanying web application to demo the proposed models which can be found at this web address\footnote{\url{https://deep-diacritization.herokuapp.com/}}.
The system uses PyTorch for implementing the neural training and inference components \cite{pytorch19}.
The LSTM cell implementation used is due to \cite{pt-rnn}\footnote{\url{https://github.com/munael/pt-rnn}}.

\subsection{Results}

\begin{table}[h]
\begin{tabular}{|l|c|c|c|c|}
\hline
\multicolumn{1}{|c|}{\multirow{2}{*}{DER/WER}} & \multicolumn{2}{c|}{Including `no diacritic'} & \multicolumn{2}{c|}{Excluding `no diacritic'} \\ \cline{2-5} 
\multicolumn{1}{|c|}{} & \multicolumn{1}{l|}{w/ case ending} & \multicolumn{1}{l|}{w/o case ending} & \multicolumn{1}{l|}{w/ case ending} & \multicolumn{1}{l|}{w/o case ending} \\ \hline
\cite{shakkala}         & 3.73\% / 11.19\%          & 2.88\% / 6.53\%         & 4.36\% / 10.89\%           & 3.33\% / 6.37\%            \\ \hline
\cite{fadel19-neural}   & 2.60\% / 7.69\%           & 2.11\% / 4.57\%         & 3.00\% / 7.39\%            & 2.42\% / 4.44\%            \\ \hline
\cite{hamza20}          & 3.39\% / 9.94\%           & 2.61\% / 5.83\%         & 3.34\% / 7.98\%            & 2.43\% / 3.98\%            \\ \hline
D2 (Ours)               & 1.85\% / 5.53\%           & 1.49\% / 3.27\%         & 2.11\% / 5.26\%            & 1.71\% / 3.15\%            \\ \hline
D3 (Ours) (@0\% hints)  & \textbf{1.83\%} / \textbf{5.34\%}  & \textbf{1.48\%} / \textbf{3.11\%}  & \textbf{2.09}\%  / \textbf{5.08\%}     & \textbf{1.69\%} / \textbf{3.00\%}   \\ \hline
\end{tabular}
\caption{Results on the Tashkeela benchmark}
\label{tashkeela-results}
\end{table}

We use the script provided by \cite{fadel19} to evaluate our results. To be consistent with prior work, we report our results in terms of both word error-rate (WER) and diacritic error-rate (DER), with and without case-endings, as well as including and excluding characters with no diacritics. Table \ref{tashkeela-results} shows our results on the Tashkeela benchmark in comparison with the more recent works. We outperform state-of-the-art by 30.56\% relative ($2.35\%$ absolute) error reduction on ``WER with case-ending''.

\begin{figure}[h]
\centering
\begin{subfigure}{.5\textwidth}
  \centering
  \includegraphics[width=1\linewidth]{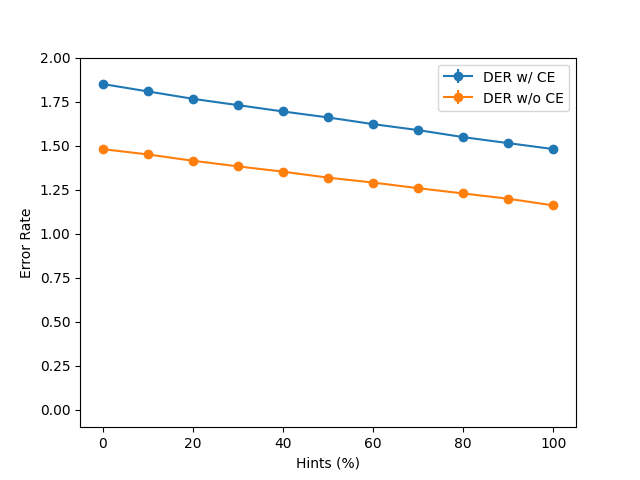}
  \caption{DER vs Percentage of Injected Hints (actual output)}
  \label{fig:pd_der_raw}
\end{subfigure}%
\begin{subfigure}{.5\textwidth}
  \centering
  \includegraphics[width=1\linewidth]{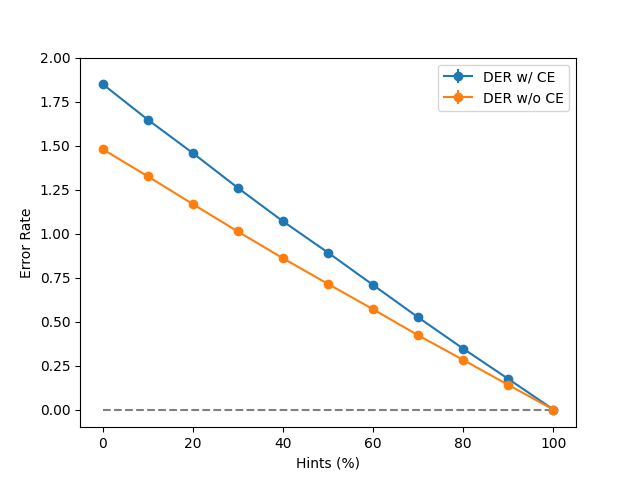}
  \caption{DER vs Percentage of Injected Hints (hints covering output)}
  \label{fig:pd_der_masked}
\end{subfigure}
\caption{Error Rate of Partially Diacritized Text}
\label{fig:pd}
\end{figure}

\begin{table}[h]
\begin{center}
\begin{tabular}{|l|r|r|r|r|c|}
\hline
\bf Method         & \bf \#Params & \bf T/epoch & \bf Convergence & \bf Inference    & \bf Full DER/WER   \\ \hline
\bf D2 -- \{Attn\} &     13.369M  & \bf 28 mins & 17 epochs       & \bf 34,996   wps & \bf 1.94\% / 5.80\%  \\
\bf Flat           & \bf 13.304M  & 121 mins    & \bf 13 epochs   &      2,466   wps &     2.20\% / 6.39\%  \\
\hline
\end{tabular}
\end{center}
\caption{Speed Comparison\protect\footnotemark}
\label{speed-table}
\end{table}
\footnotetext{All models use the same custom LSTM implementation and are run on a single Nvidia GeForce RTX 2080 Ti.}
Table \ref{speed-table} compares our plain 2-level hierarchy design (without Attention) with a ``Flat" model in task and runtime performance. The Flat model comprises a 4-layer stacked Bi-LSTM with similar implementation details as described in \ref{exp:enc-dec} for D2, including Sentence Dropout and the Voting mechanism. The Flat model sees each sentence as one sequence of characters.
\paragraph{Partially Diacritized Text}
Figure \ref{fig:pd} shows the results of DER including `no diacritic' with and without case ending when the model is supplied with partially diacritized text as input. For each character in the sentence, with some probability, we may replace the predicted output of the previous time-step with the ground truth as input to the decoder in the current step. The reported output of the previous time-step is masked to force the provided hint to be the model's ``prediction'' even if the inferred were different (see Figure \ref{fig:pd_der_masked})---in contrast to Figure \ref{fig:pd_der_raw} where the final predictions are the model's unmodified outputs. The results are averaged across five runs with different seeds (i.e. injecting the ground truth signal at different characters in the sentence). Error bars represent standard deviation. Many Arabic texts already come with some hints that can improve model performance. Here we show how a neural model could be trained to leverage that.

\section{Discussion}

\subsection{Ablation Study}

We conduct an ablation study to measure the effect of components proposed for the final model. We train and evaluate the D2 model previously detailed but with the component(s) specified removed. Table \ref{tab:ablation} shows the results after removing the sentence dropout and cross-level attention module.

\begin{table}[h]
\centering
\begin{tabular}{|l|r|r|r|r|}
\hline
\multicolumn{1}{|c|}{\multirow{2}{*}{DER/WER}} & \multicolumn{2}{c|}{Including `no-diacritic'}                              & \multicolumn{2}{c|}{Excluding `no-diacritic'}                              \\ \cline{2-5} 
\multicolumn{1}{|c|}{}                         & \multicolumn{1}{l|}{w/ case ending} & \multicolumn{1}{l|}{w/o case ending} & \multicolumn{1}{l|}{w/ case ending} & \multicolumn{1}{l|}{w/o case ending} \\ \hline
\cite{fadel19-neural}                          & 2.60\% / 7.69\%                     & 2.11\% / 4.57\%                      & 3.00\% / 7.39\%                     & 2.42\% / 4.44\%                      \\ \hline
D3 (@0\% hints)                                & \textbf{1.83\%} / \textbf{5.34\%}  & \textbf{1.48\%} / \textbf{3.11\%}  & \textbf{2.09}\%  / \textbf{5.08\%}     & \textbf{1.69\%} / \textbf{3.00\%}    \\ \hline
D2                                             & 1.85\% / 5.53\%                     & 1.49\% / 3.27\%                      & 2.11\% / 5.26\%                     & 1.71\% / 3.15\%                      \\ \hline
D2 $-$ \{Attention \ref{arch:attn}\}           & 1.94\% / 5.80\%                     & 1.58\% / 3.44\%                      & 2.23\% / 5.52\%                     & 1.80\% / 3.31\%                      \\ \hline
D2 $-$ \{SDO \ref{exp:sdo}\}                   & 1.91\% / 5.71\%                     & 1.54\% / 3.36\%                      & 2.18\% / 5.43\%                     & 1.75\% / 3.23\%                      \\ \hline
D2 $-$ \{Attn, SDO\}                           & 1.93\% / 5.78\%                     & 1.57\% / 3.45\%                      & 2.21\% / 5.49\%                     & 1.79\% / 3.32\%                      \\ \hline
\end{tabular}
\caption{Ablation Study}
\label{tab:ablation}
\end{table}

\subsection{Attention Analysis}

The cross-level attention module allows us to gauge the contribution of each word to each output diacritic. Here we examine some examples to see whether the model was able to learn such Arabic grammar rules as a human expert would use when annotating case endings. The examples presented in this section reflect patterns we have found repeated during our analysis.

\begin{figure}[!h]
\centering
\begin{subfigure}[t]{.49\textwidth}
  \centering
  \includegraphics[width=1\linewidth]{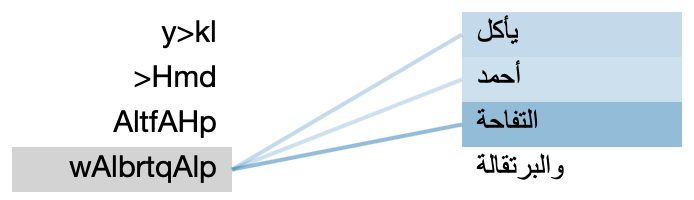}
  \caption{Ex. of case agreement by \d{h}arf-atf}
  \label{fig:ano_1}
\end{subfigure}%
\begin{subfigure}[t]{.49\textwidth}
  \centering
  \includegraphics[width=1\linewidth]{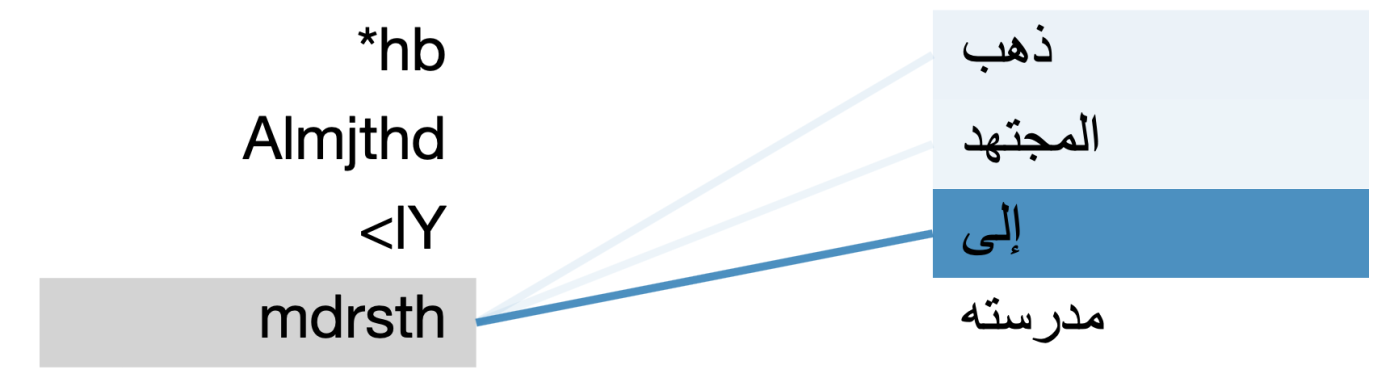}
  \caption{Ex. of CE of ism-majrūr depending on \d{h}arf-jarr}
  \label{fig:ano_2}
\end{subfigure}
\caption{Attention visualization of words correctly attending to grammatical parents.}
\label{fig:attn_ano_1}
\end{figure}

The pattern in Figure \ref{fig:ano_1} is related to the \d{h}arf-atf\footnote{\<حرف عطف>} rule (generally prepositions), which states that the word coming after it gets the same case as the main word of the phrase it is related to---the grammatical parent. We see indeed that the word coming after the (``w'') \d{h}arf-atf attends the most on the word that comes before it. This is similar to what an expert would do; look at the main word in the phrase preceding the ``w'' in order to determine the case and case-ending of what follows.

\begin{figure}[h]
\centering

\begin{subfigure}[t]{.49\textwidth}
  \centering
  \includegraphics[width=1\linewidth]{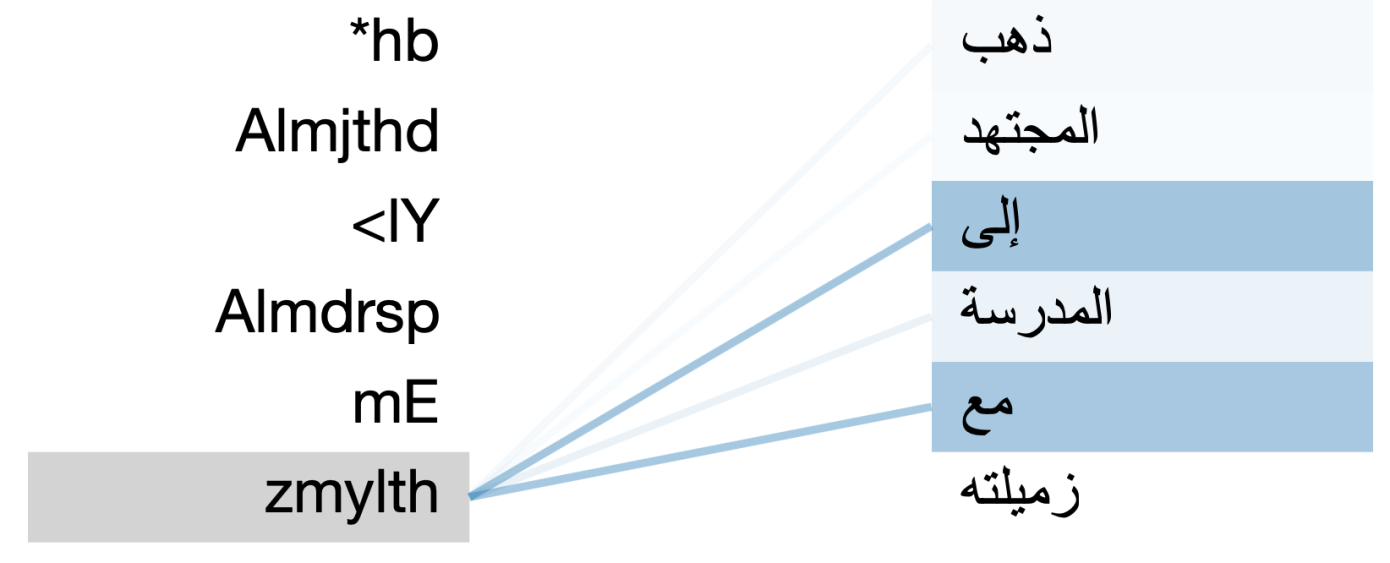}
  \caption{\emph{Self} attends on a local word with similar \emph{effect} on \emph{Self}}
  \label{fig:ano_3}
\end{subfigure} %
\begin{subfigure}[t]{.49\textwidth}
  \centering
  \includegraphics[width=1\linewidth]{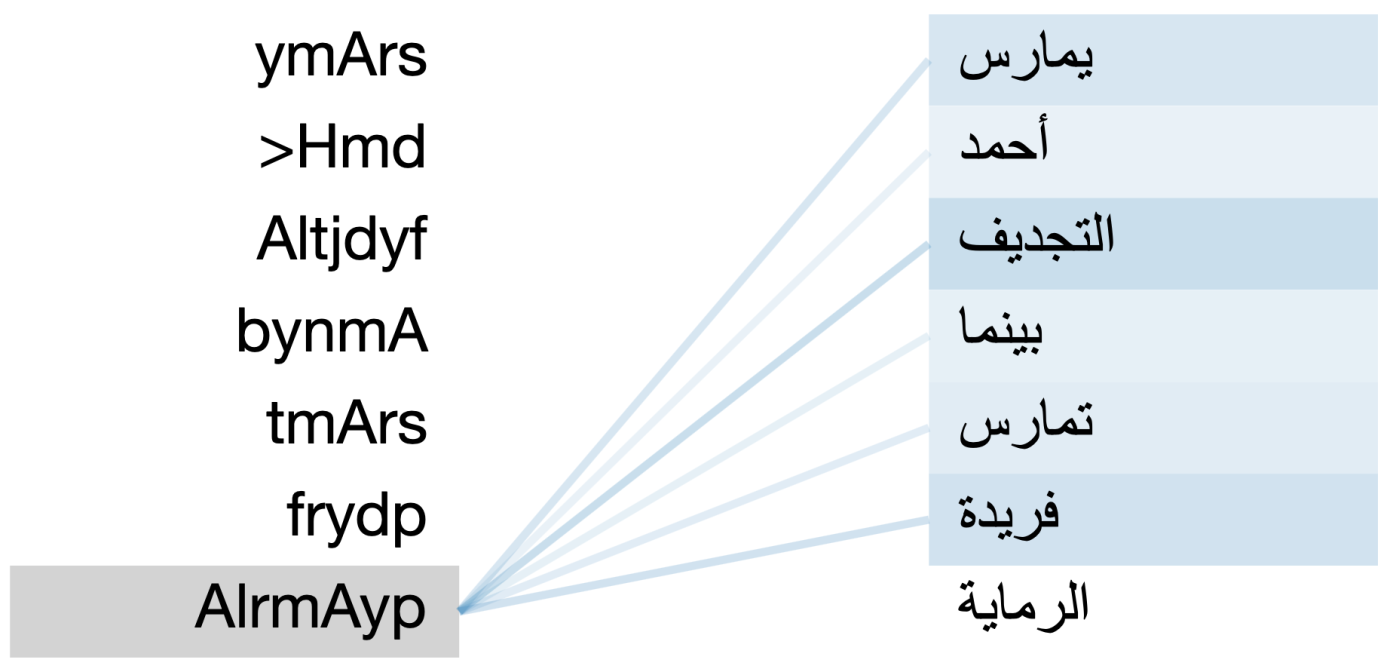}
  \caption{\emph{Self} attends on a local word with similar \emph{role} to \emph{Self}}
  \label{fig:ano_4}
\end{subfigure}
\caption{Attention visualization of confusion of grammatical parents.}
\label{fig:attn_ano_2}
\end{figure}

Figure \ref{fig:ano_2} shows another prevalent example where the word in question attends the most on the \d{h}arf-jarr\footnote{\<حرف جر>} preceding it. However, in other cases where the same rule appears twice in a segment, we found that the model may choose to attend equally or more on the components of the first occurrence rather than the occurrence the current word is actually affected by---the grammatical parent. Figure \ref{fig:ano_3} shows one example of this where the word (``zmylth'') attends equally to two words that would affect it the same (``<lY'' and ``mE''), but only the second should be affecting it. In Figure \ref{fig:ano_4} we see that the second maf'ool-bih\footnote{\<مفعول به>} (roughly an object of a verb) (``AlrmAyp'') attends heavily on the first occurrence of a mf'ool-bih in the segment (``Altjdyf''), rather than the verb that should be affecting it (``tmArs''). This behavior of attending on a previous word with a similar role suggests that the attention mechanism is aware of grammatical rules; it is able to group words with the same role together.

Generally, we found that not all sentences yield interpretable attention weights. We leave the task of comprehensively studying the extent of agreement of the learned weights with Arabic grammatical rules to future work.

\subsection{Error Analysis}

\begin{figure}[h]
\centering
\begin{subfigure}{.5\textwidth}
  \centering
  \includegraphics[width=.95\linewidth]{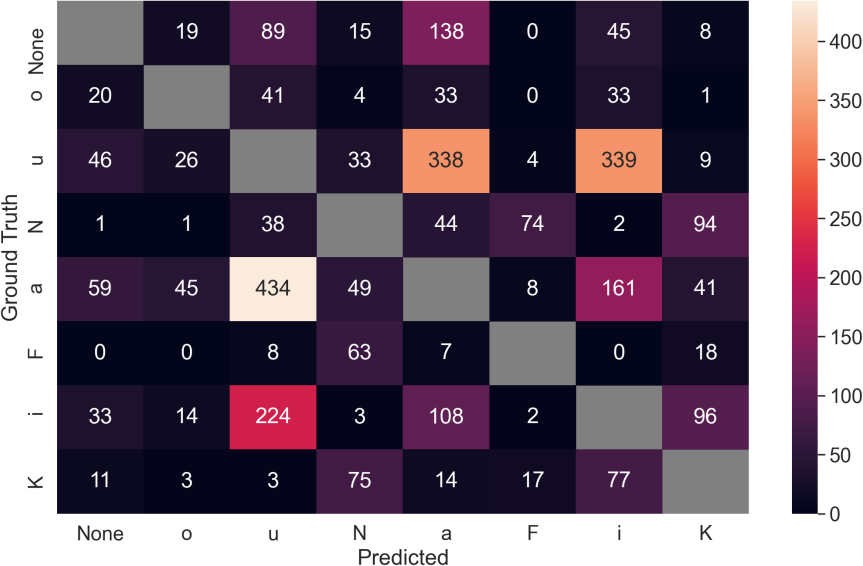}
  \caption{Confusion Matrix for Case Endings (Errors Only)}
  \label{fig:conf_ce}
\end{subfigure}%
\begin{subfigure}{.5\textwidth}
  \centering
  \includegraphics[width=.95\linewidth]{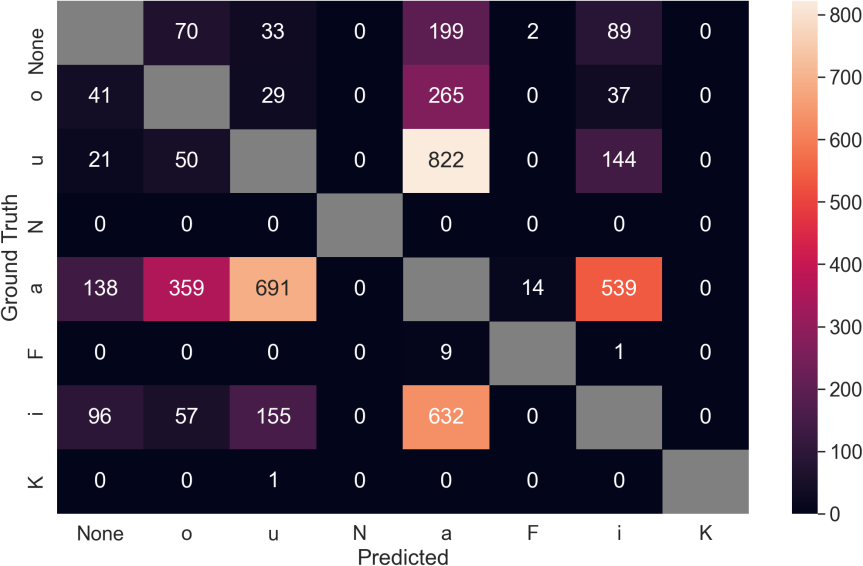}
  \caption{Confusion Matrix for Core Word Diacritics (Errors Only)}
  \label{fig:conf_core}
\end{subfigure}
\caption{Error Confusion Matrix for CE and Core Diacritics}
\label{fig:conf}
\end{figure}

Figures \ref{fig:conf_ce} and \ref{fig:conf_core} show the the confusion matrices (for visualization clarity we show errors only) of case ending and core words diacritics respectively. Many confusions are between the dammah and kasrah, and between dammah and fat\d{h}ah, and the confusion goes both ways. We analyze the errors between kasrah and dammah for case endings and try to correlate them with grammar rules. 

The first error example is related to the start of a new sentence in Arabic grammar. The word “wa” composed of one letter can either mark a conjunction (e.g. in an enumeration), or mark the start of a new sentence, based on context. An example with such confusion is in the following sentence, with the ground truth: ``wa yarud$\sim$u Ealayohi >an$\sim$a faAqida AlT$\sim$ahuwrayoni wa naHowahu layosa lahu SalaApN
<l$\sim$aA <*aA DaAqa Alowaqotu''
\footnote{\<وَيَرُدُّ عَلَيْهِ أَنَّ فَاقِدَ الطَّهُورَيْنِ وَنَحْوَهُ لَيْسَ لَهُ صَلَاةٌ إلَّا إذَا ضَاقَ الْوَقْتُ>}
. We see one confusion example where ``wa naHow\textbf{a}h'' was predicted as ``wa naHow\textbf{i}h''. Grammatically, the ``wa'' relates the next word to a ``sentence starter'' word (``faAqida'') in a case that would make \textbf{a} the correct diacritic. Instead, we observe it follows the word immediately before the ``wa'' (``AlT$\sim$ahuwr\textbf{ayoni}''), which is indeed in a grammatical case that would make \textbf{i} the correct diacritic for ``naHow\{\textbf{a}/i\}h'', were it the correct grammatical parent of this ``wa''.

The second example is related to the use of punctuation marks that signal an abrupt start of a new sentence or an end of one with unique context that may not be easily learnt. In the following sentence with the ground truth: ``kaqaA\}ilK : AloHar$\sim$\textbf{u} >awo Alobarod\textbf{u} Al\$$\sim$adiyd\textbf{u}''
\footnote{\<كَقَائِلٍ: الْحَرُّ أَوْ الْبَرْدُ الشَّدِيدُ...>} 
was predicted as ``kaqaA\}ilK : AloHar$\sim$\textbf{i} >awo Alobarod\textbf{i} Al\$$\sim$adiyd\textbf{i}...''. The mark “:” here denotes the start of a new sentence, by convention, as we start a quotation. In speech, this would manifest as a brief pause or change in tone. Without “:”, the word would be an ism-majrūr that takes the kasrah \d{h}arakah (\textbf{i}) in this position, which the model mistakenly outputs. But because it starts a new sentence, the correct diacritic is a dammah \d{h}arakah (\textbf{u}). Further, the predictions for the words following ``>awo'' behave grammatically by following the case of the parent word (``AloHar$\sim$\textbf{i}'') (according to ``wa''), but are incorrect because the error has propagated.

One other type of errors is related to inconsistencies in the corpus---the same word with the same role in the sentence is not diacritized the same way across the dataset. For instance, the word ``<lY'', which is the second top word that causes a core word error as shown in Table \ref{top-error}, appears multiple times in different forms: ``<ilY'', ``<lY'', and ``<ilaY''---all correct. There are other examples that show the need to clean the dataset (at least the test set) to evaluate the published models properly.

\begin{table}[h]
\begin{tabular}{|r|c|c|c|c|c|c|c|c|c|c|}
\hline
\textbf{Rank}                               & \textbf{1} & \textbf{2} & \textbf{3} & \textbf{4} & \textbf{5} & \textbf{6} & \textbf{7} & \textbf{8} & \textbf{9} & \textbf{10} \\ \hline
\textbf{Top Words w/ Wrong CE Diacritics}   & \<غير>     & \<عن>      & \<كل>      & \<بن>      & \<قلت>     & \<مثل>     &  \<ثم>     & \<يوم>     &  \<و>      &     \<من>   \\ \hline
\textbf{Top Words w/ Wrong Core Diacritics} & \<اللّه>    & \<إلى>     & \<ذكر>     & \<من>      &  \<إلا>     &  \<علم>    &  \<إن>     & \<قبل>     & \<وسلّم>    &   \<قوله>   \\ \hline
\end{tabular}
\caption{Top 10 Words with CE and Core Word Errors}
\label{top-error}
\end{table}

Looking at the top words that yield confusions in both core words and case ending diacritics, we find a notable intersection between Table \ref{top-error} and the most frequent tokens in the Tashkeela corpus with an average max-normalized frequency of $0.24$ ($0.24$ as frequent as the most frequent word).

\section{Conclusion}
In this work, we presented a novel architecture that outperforms previously published results on the Tashkeela Arabic diacritization benchmark. Future work may include:
\begin{itemize}
    \setlength\itemsep{-0.3em}
    \item Replacing the word- and character- level Bi-LSTM encoders with transformer-based encoders.
    \item Using byte-pair-encoding (BPE) \cite{bpe-16} to better handle suffixes and prefixes as Arabic is a moderately fusional language.
    \item Investigating more efficient use of injected hints to improve performance.
    \item Training/Evaluating this design/model on Modern Standard Arabic and dialectical benchmarks.
    \item Cleaning the testset of Tashkeela to remove any inconsistencies as described in the error analysis.
    \item Finally, achieving more interpretable attention weights through multi-task training, training on larger datasets, or otherwise.
\end{itemize}

\section*{Acknowledgements}
We offer special thanks to \textit{Khaled Essam}, as well as \textit{Mohamed Afify} and \textit{Ahmed Tawfik} of \textit{Microsoft EGDC}, for many helpful discussions, suggestions and comments on the paper.

\bibliographystyle{acl}
\bibliography{references}

\begin{thebibliography}{}

\bibitem[\protect\citename{Abbad and Xiong}2020]{hamza20}
Hamza Abbad and Shengwu Xiong.
\newblock 2020.
\newblock Multi-components system for automatic arabic diacritization.
\newblock In Joemon~M. Jose, Emine Yilmaz, Jo{\~a}o Magalh{\~a}es, Pablo
  Castells, Nicola Ferro, M{\'a}rio~J. Silva, and Fl{\'a}vio Martins, editors,
  {\em Advances in Information Retrieval}, pages 341--355, Cham. Springer
  International Publishing.

\bibitem[\protect\citename{Azmi}2013]{azmi_2013}
Aqil Azmi.
\newblock 2013.
\newblock A survey of automatic arabic diacritization techniques.
\newblock {\em Natural Language Engineering}, 21, 10.

\bibitem[\protect\citename{Barqawi}2017]{shakkala}
Zerrouki Barqawi.
\newblock 2017.
\newblock Shakkala, {A}rabic text vocalization.
\newblock \url{https://github.com/Barqawiz/Shakkala}.

\bibitem[\protect\citename{Belinkov and Glass}2015]{belinkov-glass-2015-arabic}
Yonatan Belinkov and James Glass.
\newblock 2015.
\newblock {A}rabic diacritization with recurrent neural networks.
\newblock In {\em Proceedings of the 2015 Conference on Empirical Methods in
  Natural Language Processing}, pages 2281--2285, Lisbon, Portugal, September.
  Association for Computational Linguistics.

\bibitem[\protect\citename{Bojanowski \bgroup et al.\egroup }2017]{fastText17}
Piotr Bojanowski, Edouard Grave, Armand Joulin, and Tomas Mikolov.
\newblock 2017.
\newblock Enriching word vectors with subword information.
\newblock {\em Transactions of the Association for Computational Linguistics},
  5:135--146.

\bibitem[\protect\citename{Chung \bgroup et al.\egroup }2014]{gru14}
Junyoung Chung, {\c{C}}aglar G{\"{u}}l{\c{c}}ehre, KyungHyun Cho, and Yoshua
  Bengio.
\newblock 2014.
\newblock Empirical evaluation of gated recurrent neural networks on sequence
  modeling.
\newblock {\em CoRR}, abs/1412.3555.

\bibitem[\protect\citename{Darwish \bgroup et al.\egroup }2017]{darwish17}
Kareem Darwish, Hamdy Mubarak, and Ahmed Abdelali.
\newblock 2017.
\newblock {A}rabic diacritization: Stats, rules, and hacks.
\newblock In {\em Proceedings of the Third {A}rabic Natural Language Processing
  Workshop}, pages 9--17, Valencia, Spain, April. Association for Computational
  Linguistics.

\bibitem[\protect\citename{{E}l{N}okrashy}2020]{pt-rnn}
{M}uhammad~N. {E}l{N}okrashy.
\newblock 2020.
\newblock {Extensible RNN cells for PyTorch}.
\newblock \url{https://github.com/munael/pt-rnn}.

\bibitem[\protect\citename{Elshafei \bgroup et al.\egroup }2006]{elshafei06}
Moustafa Elshafei, Husni Al-Muhtaseb, and Mansour Alghamdi.
\newblock 2006.
\newblock Statistical methods for automatic diacritization of arabic text.
\newblock {\em The Saudi 18th National Computer Conference. Riyadh},
  18:301--306, 01.

\bibitem[\protect\citename{{Fadel} \bgroup et al.\egroup }2019a]{fadel19}
Ali {Fadel}, Ibraheem {Tuffaha}, Bara' {Al-Jawarneh}, and Mahmoud {Al-Ayyoub}.
\newblock 2019a.
\newblock Arabic text diacritization using deep neural networks.
\newblock In {\em 2019 2nd International Conference on Computer Applications
  Information Security (ICCAIS)}, pages 1--7, May.

\bibitem[\protect\citename{Fadel \bgroup et al.\egroup }2019b]{fadel19-neural}
Ali Fadel, Ibraheem Tuffaha, Bara{'} Al-Jawarneh, and Mahmoud Al-Ayyoub.
\newblock 2019b.
\newblock Neural {A}rabic text diacritization: State of the art results and a
  novel approach for machine translation.
\newblock In {\em Proceedings of the 6th Workshop on Asian Translation}, pages
  215--225, Hong Kong, China, November. Association for Computational
  Linguistics.

\bibitem[\protect\citename{Gal and Ghahramani}2016]{rec-dropout}
Yarin Gal and Zoubin Ghahramani.
\newblock 2016.
\newblock A theoretically grounded application of dropout in recurrent neural
  networks.
\newblock In D.~D. Lee, M.~Sugiyama, U.~V. Luxburg, I.~Guyon, and R.~Garnett,
  editors, {\em Advances in Neural Information Processing Systems 29}, pages
  1019--1027. Curran Associates, Inc.

\bibitem[\protect\citename{Graves \bgroup et al.\egroup }2005]{graves05}
Alex Graves, Santiago Fern{\'a}ndez, and J{\"u}rgen Schmidhuber.
\newblock 2005.
\newblock {Bidirectional LSTM Networks for Improved Phoneme Classification and
  Recognition}.
\newblock In W{\l}odzis{\l}aw Duch, Janusz Kacprzyk, Erkki Oja, and S{\l}awomir
  Zadro{\.{z}}ny, editors, {\em Artificial Neural Networks: Formal Models and
  Their Applications -- ICANN 2005}, pages 799--804, Berlin, Heidelberg.
  Springer Berlin Heidelberg.

\bibitem[\protect\citename{Habash \bgroup et al.\egroup }2009]{MADA}
Nizar Habash, Owen Rambow, and Ryan Roth.
\newblock 2009.
\newblock {MADA+TOKAN: A toolkit for Arabic tokenization, diacritization,
  morphological disambiguation, POS tagging, stemming and lemmatization}.
\newblock {\em Proceedings of the 2nd International Conference on Arabic
  Language Resources and Tools (MEDAR)}, 01.

\bibitem[\protect\citename{Kingma and Ba}2014]{adam14}
Diederik~P. Kingma and Jimmy Ba.
\newblock 2014.
\newblock Adam: A method for stochastic optimization.
\newblock cite arxiv:1412.6980Comment: Published as a conference paper at the
  3rd International Conference for Learning Representations, San Diego, 2015.

\bibitem[\protect\citename{Mubarak \bgroup et al.\egroup
  }2019]{mubarak19-highly}
Hamdy Mubarak, Ahmed Abdelali, Hassan Sajjad, Younes Samih, and Kareem Darwish.
\newblock 2019.
\newblock Highly effective {A}rabic diacritization using sequence to sequence
  modeling.
\newblock In {\em Proceedings of the 2019 Conference of the North {A}merican
  Chapter of the Association for Computational Linguistics: Human Language
  Technologies, Volume 1 (Long and Short Papers)}, pages 2390--2395,
  Minneapolis, Minnesota, June. Association for Computational Linguistics.

\bibitem[\protect\citename{Nelken and Shieber}2005]{shieber05}
Rani Nelken and Stuart~M. Shieber.
\newblock 2005.
\newblock {A}rabic diacritization using weighted finite-state transducers.
\newblock In {\em Proceedings of the {ACL} Workshop on Computational Approaches
  to {S}emitic Languages}, pages 79--86, Ann Arbor, Michigan, June. Association
  for Computational Linguistics.

\bibitem[\protect\citename{Pasha \bgroup et al.\egroup }2014]{pasha14-madamira}
Arfath Pasha, Mohamed Al-Badrashiny, Mona Diab, Ahmed El~Kholy, Ramy Eskander,
  Nizar Habash, Manoj Pooleery, Owen Rambow, and Ryan Roth.
\newblock 2014.
\newblock {MADAMIRA}: A fast, comprehensive tool for morphological analysis and
  disambiguation of {A}rabic.
\newblock In {\em Proceedings of the Ninth International Conference on Language
  Resources and Evaluation ({LREC}'14)}, pages 1094--1101, Reykjavik, Iceland,
  May. European Language Resources Association (ELRA).

\bibitem[\protect\citename{Paszke \bgroup et al.\egroup }2019]{pytorch19}
Adam Paszke, Sam Gross, Francisco Massa, Adam Lerer, James Bradbury, Gregory
  Chanan, Trevor Killeen, Zeming Lin, Natalia Gimelshein, Luca Antiga, Alban
  Desmaison, Andreas Kopf, Edward Yang, Zachary DeVito, Martin Raison, Alykhan
  Tejani, Sasank Chilamkurthy, Benoit Steiner, Lu~Fang, Junjie Bai, and Soumith
  Chintala.
\newblock 2019.
\newblock Pytorch: An imperative style, high-performance deep learning library.
\newblock In H.~Wallach, H.~Larochelle, A.~Beygelzimer, F.~d\textquotesingle
  Alch\'{e}-Buc, E.~Fox, and R.~Garnett, editors, {\em Advances in Neural
  Information Processing Systems 32}, pages 8024--8035. Curran Associates, Inc.

\bibitem[\protect\citename{Sennrich \bgroup et al.\egroup }2016]{bpe-16}
Rico Sennrich, Barry Haddow, and Alexandra Birch.
\newblock 2016.
\newblock Neural machine translation of rare words with subword units.
\newblock In {\em Proceedings of the 54th Annual Meeting of the Association for
  Computational Linguistics (Volume 1: Long Papers)}, pages 1715--1725, Berlin,
  Germany, August. Association for Computational Linguistics.

\bibitem[\protect\citename{Srivastava \bgroup et al.\egroup }2014]{dropout}
Nitish Srivastava, Geoffrey Hinton, Alex Krizhevsky, Ilya Sutskever, and Ruslan
  Salakhutdinov.
\newblock 2014.
\newblock Dropout: A simple way to prevent neural networks from overfitting.
\newblock {\em Journal of Machine Learning Research}, 15(56):1929--1958.

\bibitem[\protect\citename{Tinsley and Board}2013]{tinsley_board}
Teresa Tinsley and Kathryn Board.
\newblock 2013.
\newblock {\em {L}anguages for the {F}uture}.
\newblock British Council.

\bibitem[\protect\citename{Vaswani \bgroup et al.\egroup }2017]{attention}
Ashish Vaswani, Noam Shazeer, Niki Parmar, Jakob Uszkoreit, Llion Jones,
  Aidan~N. Gomez, Lukasz Kaiser, and Illia Polosukhin.
\newblock 2017.
\newblock Attention is all you need.
\newblock {\em CoRR}, abs/1706.03762.

\bibitem[\protect\citename{Zalmout and Habash}2020]{joint2020}
Nasser Zalmout and Nizar Habash.
\newblock 2020.
\newblock Joint diacritization, lemmatization, normalization, and fine-grained
  morphological tagging.
\newblock In {\em Proceedings of the 58th Annual Meeting of the Association for
  Computational Linguistics}, pages 8297--8307, Online, July. Association for
  Computational Linguistics.

\bibitem[\protect\citename{Zerrouki and Balla}2017]{tashkeela17}
Taha Zerrouki and Amar Balla.
\newblock 2017.
\newblock Tashkeela: Novel corpus of arabic vocalized texts, data for
  auto-diacritization systems.
\newblock {\em Data in Brief}, 11:147 -- 151.

\bibitem[\protect\citename{Zitouni and Sarikaya}2009]{zitouni09}
Imed Zitouni and Ruhi Sarikaya.
\newblock 2009.
\newblock Arabic diacritic restoration approach based on maximum entropy
  models.
\newblock {\em Computer Speech \& Language}, 23:257--276, 07.

\bibitem[\protect\citename{Zitouni \bgroup et al.\egroup }2006]{zitouni06}
Imed Zitouni, Jeffrey~S. Sorensen, and Ruhi Sarikaya.
\newblock 2006.
\newblock Maximum entropy based restoration of {A}rabic diacritics.
\newblock In {\em Proceedings of the 21st International Conference on
  Computational Linguistics and 44th Annual Meeting of the Association for
  Computational Linguistics}, pages 577--584, Sydney, Australia, July.
  Association for Computational Linguistics.

\end{thebibliography}

\end{document}